\documentclass[11pt,a4paper]{article}

\usepackage[utf8]{inputenc}
\usepackage[T1]{fontenc}
\usepackage{times}
\usepackage[margin=2.5cm]{geometry}
\usepackage{graphicx}
\usepackage{hyperref}
\usepackage{xcolor}
\usepackage{tikz}
\usetikzlibrary{shapes.geometric, arrows.meta, positioning, fit, backgrounds}
\usepackage{amsmath}
\usepackage{amssymb}
\usepackage{booktabs}
\usepackage{natbib}
\usepackage{listings}
\usepackage{algorithm}
\usepackage{algpseudocode}
\usepackage{authblk}

\hypersetup{
    colorlinks=true,
    linkcolor=blue,
    citecolor=blue,
    urlcolor=blue
}

\tikzstyle{startstop} = [rectangle, rounded corners, minimum width=2.5cm, minimum height=0.8cm, text centered, draw=black, fill=blue!20]
\tikzstyle{process} = [rectangle, minimum width=3cm, minimum height=0.8cm, text centered, draw=black, fill=orange!20]
\tikzstyle{decision} = [diamond, minimum width=2.5cm, minimum height=0.8cm, text centered, draw=black, fill=green!20, aspect=2]
\tikzstyle{io} = [trapezium, trapezium left angle=70, trapezium right angle=110, minimum width=2cm, minimum height=0.8cm, text centered, draw=black, fill=yellow!20]
\tikzstyle{arrow} = [thick,->,>=stealth]

\title{CheckIfExist: Detecting Citation Hallucinations in the Era of AI-Generated Content}

\author{Diletta Abbonato\thanks{diletta.abbonato@unito.it}}
\affil{Department of Culture, Politics and Society \\ University of Turin}

\date{}

\begin{document}

\maketitle

\begin{abstract}
The proliferation of large language models (LLMs) in academic workflows has introduced unprecedented challenges to bibliographic integrity, particularly through reference hallucination---the generation of plausible but non-existent citations. Recent investigations have documented the presence of AI-hallucinated citations even in papers accepted at premier machine learning conferences such as NeurIPS and ICLR, underscoring the urgency of automated verification mechanisms. This paper presents \textsc{CheckIfExist}, an open-source web-based tool designed to provide immediate verification of bibliographic references through multi-source validation against CrossRef, Semantic Scholar, and OpenAlex scholarly databases. While existing reference management tools offer bibliographic organization capabilities, they do not provide real-time validation of citation authenticity. Commercial hallucination detection services, though increasingly available, often impose restrictive usage limits on free tiers or require substantial subscription fees. The proposed tool fills this gap by employing a cascading validation architecture with string similarity algorithms to compute multi-dimensional match confidence scores, delivering instant feedback on reference authenticity. The system supports both single-reference verification and batch processing of BibTeX entries through a unified interface, returning validated APA citations and exportable BibTeX records within seconds. 

\vspace{0.3cm}
\end{abstract}

\section{Introduction}
This paper introduces CheckIfExist, an open-source tool for real-time validation of bibliographic references against multiple scholarly databases, designed to address the growing problem of citation hallucinations in AI-generated academic content.\footnote{Available at \url{https://zabbonat.github.io/References-Validation/}}

The integrity of scholarly communication fundamentally depends upon the accuracy and verifiability of bibliographic references. Citations serve multiple critical functions within the scientific ecosystem: they acknowledge intellectual debt, enable readers to trace the provenance of ideas, facilitate reproducibility through access to source materials, and form the basis for quantitative assessments of research impact \citep{garfield1972citation, merton1973sociology}. The emergence of citation analysis as a cornerstone of scientometrics---from journal impact factors to h-indices---has further elevated the importance of citation accuracy, as erroneous references can propagate through citation networks and distort bibliometric indicators \citep{macroberts1996problems, taskin2025frozen}.

The rapid adoption of large language models in academic workflows has introduced a novel threat to bibliographic integrity: reference hallucination. These AI systems, trained on vast corpora of text, can generate citations that appear superficially plausible---featuring realistic author names, journal titles, and publication years---yet correspond to no actual published work \citep{alkaissi2023artificial, athaluri2023exploring}. The phenomenon is particularly insidious because hallucinated references often exhibit the stylistic conventions of legitimate citations, rendering manual detection cognitively demanding and time-consuming \citep{dunford2024automated}. The severity of this problem has been starkly illustrated by recent investigations at premier AI research conferences. An analysis of over 4,000 research papers accepted and presented at NeurIPS 2025 uncovered more than 100 AI-hallucinated citations across at least 53 papers---references that had passed peer review by three or more reviewers and entered the official conference proceedings \citep{goldman2026neurips}. These hallucinations ranged from fully fabricated citations with non-existent authors to subtle alterations of real papers, such as expanding author initials into guessed first names or paraphrasing titles. Similar findings emerged from papers under review at ICLR, where 50 hallucinated citations were identified before acceptance. Given that NeurIPS 2025 had an acceptance rate of 24.52\% from over 21,500 submissions, the presence of fabricated references in accepted papers represents a significant breach of scholarly standards.

The scholarly community has developed sophisticated tools for bibliographic management. Reference managers such as Zotero, Mendeley, EndNote, and JabRef provide comprehensive functionality for organizing, storing, and formatting citations \citep{kratochvil2017comparison}. These tools excel at importing metadata from databases, synchronizing libraries across devices, generating formatted bibliographies, and integrating with word processors. However, these systems are fundamentally designed for organization rather than validation. When a researcher adds a reference---whether manually entered, imported from a PDF, or copied from an AI-generated text---these tools store the provided metadata without verifying its authenticity. A hallucinated reference with a fabricated DOI, non-existent journal, or invented author will be catalogued alongside legitimate citations without any indication of its spurious nature. Similarly, bibliographic databases such as Web of Science, Scopus, and Google Scholar provide search functionality that can be used to verify references, but this requires manual querying of each citation---a process that scales poorly and interrupts the writing workflow. Commercial services for hallucination detection have emerged in response to the AI-generated content problem, but these typically operate under restrictive freemium models: free tiers may limit users to a small number of verifications per day or month, while comprehensive access requires institutional or individual subscriptions that may be prohibitive for researchers in resource-constrained environments \citep{zhu2025llm}.

The tool presented in this paper addresses precisely this gap: it provides instant verification of reference authenticity through real-time queries to multiple scholarly databases---CrossRef, Semantic Scholar, and OpenAlex---returning validation results within seconds rather than requiring researchers to interrupt their work for manual database searches or navigate paywall barriers.

\section{Background and Motivation}

The architecture of modern science rests upon an implicit social contract: citations function as verifiable claims that connect new knowledge to its intellectual antecedents. This system operates as a distributed verification mechanism wherein each reference constitutes a traceable node in the broader network of scholarly communication \citep{merton1973sociology}. Yet this infrastructure has always been vulnerable to degradation. Studies examining citation accuracy across disciplines have documented error rates ranging from 25\% to 54\% of references containing at least one inaccuracy \citep{siebers2000accuracy, wager2008effects}, with consequences that extend well beyond individual papers. Citation errors generate negative externalities throughout the knowledge network: they misdirect subsequent researchers, waste resources when scholars attempt to locate non-existent sources, and---most critically---undermine the cumulative and self-correcting nature of scientific inquiry \citep{simkin2003read, jung2025citation}. The propagation dynamics are particularly concerning, as mistakes in highly-cited works replicate across the literature through a process of ``citation mutation'' \citep{wright2008ombudsman}, suggesting that the reliability of the scholarly record exhibits path-dependent vulnerabilities.

The deployment of large language models in academic workflows has fundamentally altered the calculus of this problem. Traditional citation errors arise primarily from transcription mistakes---an essentially random process with error rates constrained by human attention costs. LLM hallucinations represent a categorically different phenomenon: systematic fabrications emerging from statistical regularities in language generation rather than from engagement with actual sources \citep{ji2023survey}. These systems exploit the very conventions that make citations trustworthy, generating references that exhibit sophisticated verisimilitude---plausible author names in relevant fields, journals that publish on adjacent topics, titles aligned with expected research trajectories. Empirical investigations have documented hallucination rates ranging from 6\% to over 30\% depending on model, prompting strategy, and domain \citep{agrawal2024language}, with fabrications often taking ``chimeric'' forms that blend elements from multiple real papers into single non-existent entries \citep{dunford2024automated}.

The economic implications merit careful consideration. The marginal cost of generating plausible-looking citations has collapsed to near zero, while the verification costs borne by readers, reviewers, and editors remain substantial. This asymmetry creates a classic market failure: the private benefits of rapid content production accrue to authors, while the costs of degraded information quality are distributed across the entire scholarly community. Evidence already suggests that AI-generated content with undisclosed LLM usage is appearing in peer-reviewed journals indexed in Web of Science and Scopus, with many such papers subsequently accumulating citations \citep{strzelecki2024chatgpt}---effectively laundering fabricated references into the legitimate scholarly record. The stochastic and scalable nature of these hallucinations necessitates verification mechanisms that can operate at comparable speed and scale, shifting the burden from costly human scrutiny to automated validation systems capable of querying authoritative metadata sources in real time.
\section{System Architecture}

The proposed system is implemented as a web-based application using React with TypeScript, ensuring cross-platform compatibility and eliminating installation barriers. The application architecture separates concerns into four primary modules: an input preprocessing service for LaTeX command filtering, a BibTeX parsing service for structured input processing, a multi-source search service implementing the cascading verification logic, and a presentation layer for result visualization and export functionality.

The tool implements a cascading validation architecture that queries three complementary scholarly databases. CrossRef serves as the primary source, providing access to metadata for over 140 million scholarly works registered through the Digital Object Identifier system and representing the most comprehensive source of standardized bibliographic metadata from thousands of publishers worldwide \citep{hendricks2020crossref}. Semantic Scholar, maintained by the Allen Institute for AI, functions as a fallback source indexing over 200 million academic papers with enhanced metadata including author disambiguation and citation context \citep{ammar2018semantic}. OpenAlex completes the validation cascade as a fully open scholarly metadata source aggregating data from CrossRef, PubMed, ORCID, arXiv, and other repositories, providing comprehensive coverage particularly for open-access and non-English publications \citep{priem2022openalex, cespedes2025openalex}. This multi-source approach addresses a fundamental limitation of single-database validation: no individual scholarly database achieves complete coverage of the academic literature. CrossRef excels for DOI-registered works but may lack coverage of preprints, regional journals, or older publications; Semantic Scholar provides strong coverage of computer science and biomedical literature with enhanced disambiguation features; OpenAlex offers the broadest coverage by aggregating multiple sources, though metadata quality varies \citep{cespedes2025openalex}. By cascading through these sources and cross-validating author information, the system achieves higher recall than any single-source approach while maintaining precision through multi-source confirmation.

A critical preprocessing step addresses a common source of validation errors: the presence of LaTeX formatting commands within copied reference text. Researchers frequently paste references directly from LaTeX documents, which may contain commands such as \texttt{\textbackslash vspace}, \texttt{\textbackslash hspace}, \texttt{\textbackslash textit}, or custom macros that interfere with parsing and search operations. The input preprocessing module automatically filters these commands, extracting clean bibliographic text for subsequent processing and enabling seamless validation of references copied directly from manuscript source files without manual cleaning.

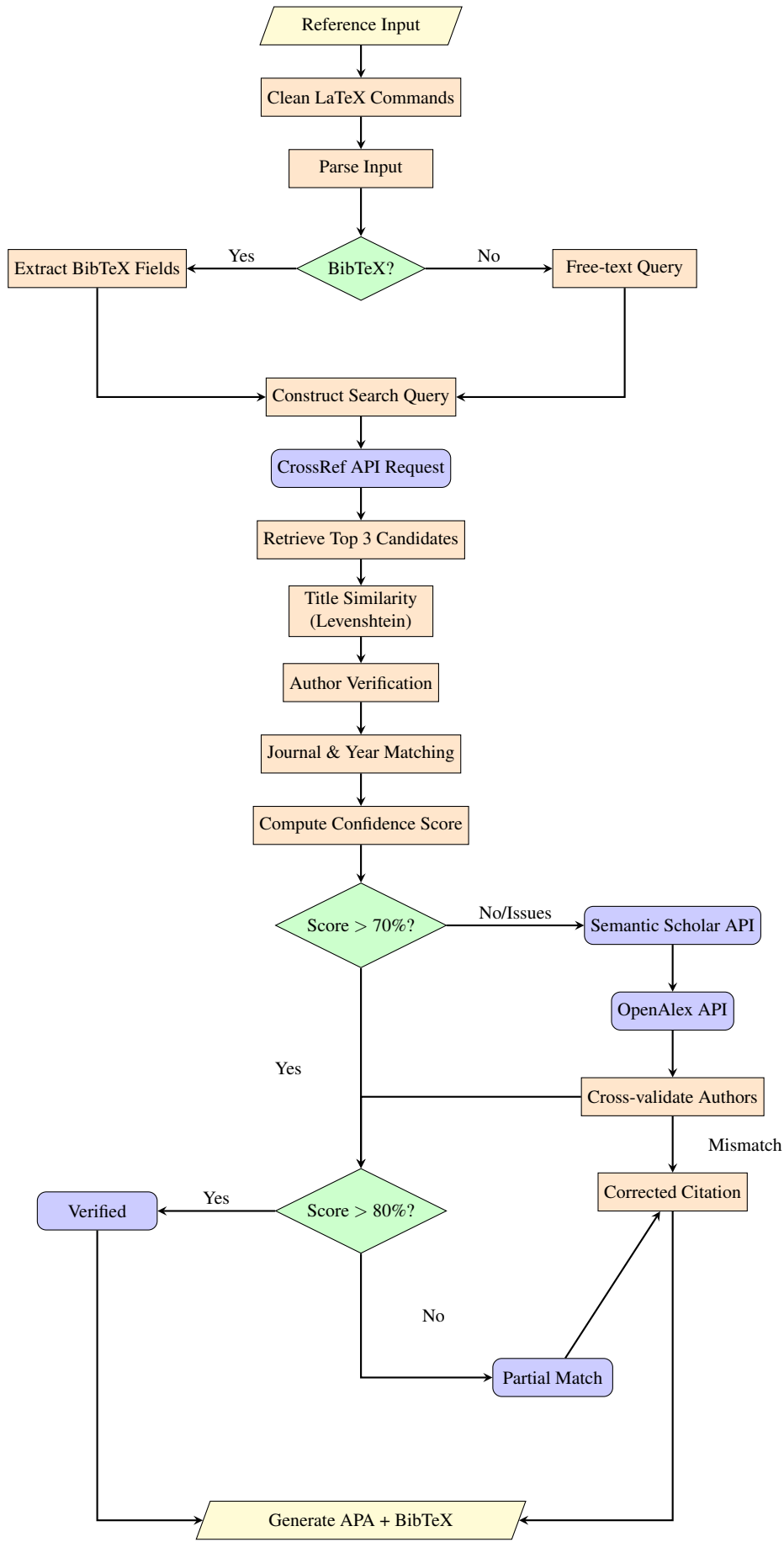
\begin{figure}[htpb!]
\centering
\begin{tikzpicture}[node distance=1.5cm, scale=0.75, transform shape,
    every node/.style={minimum width=3cm, align=center}]

\node (input) [io] {Reference Input};
\node (clean) [process, below of=input] {Clean LaTeX Commands};
\node (parse) [process, below of=clean] {Parse Input};
\node (detect) [decision, below of=parse, yshift=-0.6cm] {BibTeX?};

\node (bibtex) [process, left of=detect, xshift=-4cm] {Extract BibTeX Fields};
\node (freetext) [process, right of=detect, xshift=4cm] {Free-text Query};

\node (query) [process, below of=detect, yshift=-1.2cm] {Construct Search Query};

\node (crossref) [startstop, below of=query] {CrossRef API Request};

\node (candidates) [process, below of=crossref] {Retrieve Top 3 Candidates};

\node (title) [process, below of=candidates] {Title Similarity\\(Levenshtein)};
\node (author) [process, below of=title] {Author Verification};
\node (meta) [process, below of=author] {Journal \& Year Matching};

\node (score) [process, below of=meta] {Compute Confidence Score};

\node (threshold1) [decision, below of=score, yshift=-0.6cm] {Score $>$ 70\%?};

\node (semantic) [startstop, right of=threshold1, xshift=5cm] {Semantic Scholar API};
\node (openalex) [startstop, below of=semantic, yshift=-0.3cm] {OpenAlex API};
\node (fallback) [process, below of=openalex, yshift=-0.3cm] {Cross-validate Authors};

\node (corrected) [process, below of=fallback, yshift=-0.5cm] {Corrected Citation};

\node (threshold2) [decision, below of=threshold1, yshift=-4.5cm] {Score $>$ 80\%?};

\node (verified) [startstop, left of=threshold2, xshift=-4cm] {Verified};
\node (partial) [startstop, below of=threshold2, yshift=-2cm, xshift=4cm] {Partial Match};

\node (output) [io, below of=threshold2, yshift=-5cm] {Generate APA + BibTeX};

\draw [arrow] (input) -- (clean);
\draw [arrow] (clean) -- (parse);
\draw [arrow] (parse) -- (detect);
\draw [arrow] (detect) -- node[above] {Yes} (bibtex);
\draw [arrow] (detect) -- node[above] {No} (freetext);
\draw [arrow] (bibtex) |- (query);
\draw [arrow] (freetext) |- (query);
\draw [arrow] (query) -- (crossref);
\draw [arrow] (crossref) -- (candidates);
\draw [arrow] (candidates) -- (title);
\draw [arrow] (title) -- (author);
\draw [arrow] (author) -- (meta);
\draw [arrow] (meta) -- (score);
\draw [arrow] (score) -- (threshold1);

\draw [arrow] (threshold1) -- node[above] {No/Issues} (semantic);
\draw [arrow] (semantic) -- (openalex);
\draw [arrow] (openalex) -- (fallback);
\draw [arrow] (fallback) -- node[right] {Mismatch} (corrected);
\draw [arrow] (fallback.west) -| (threshold2.north);

\draw [arrow] (threshold1.south) -- node[left] {Yes} (threshold2.north);

\draw [arrow] (threshold2) -- node[above] {Yes} (verified);
\draw [arrow] (threshold2.south) |- node[near start, right] {No} (partial.west);
\draw [arrow] (verified) |- (output);
\draw [arrow] (partial) -- (corrected);
\draw [arrow] (corrected.south) |- (output.east);

\end{tikzpicture}
\caption{Multi-source reference validation workflow with fallback to Semantic Scholar and OpenAlex APIs.}
\label{fig:pipeline}
\end{figure}

Figure~\ref{fig:pipeline} illustrates the complete verification pipeline from input to output generation, including the multi-source fallback mechanism.

The verification process employs a multi-stage matching algorithm with cascading fallback designed to maximize both precision and recall across heterogeneous input formats and database coverage gaps. Algorithm~\ref{alg:verification} presents the pseudocode for the core verification procedure including the multi-source validation logic.

\begin{algorithm}[ht]
\caption{Multi-Source Reference Verification Algorithm}
\label{alg:verification}
\begin{algorithmic}[1]
\Require Reference query $q$, optional expected metadata $E$
\Ensure Verification result with confidence score and validated metadata

\State $q \gets \textsc{FilterLaTeXCommands}(q)$ \Comment{Remove \texttt{\textbackslash vspace}, \texttt{\textbackslash hspace}, etc.}
\State $candidates \gets \textsc{CrossRefQuery}(q, rows=3)$
\If{$candidates = \emptyset$}
    \State $candidates \gets \textsc{SemanticScholarQuery}(q)$
\EndIf

\State $bestMatch, bestScore \gets \textsc{EvaluateCandidates}(q, candidates)$
\State $issues \gets \textsc{DetectIssues}(q, bestMatch)$

\If{$bestScore < 70$ \textbf{or} $|issues| > 0$}
    \State \Comment{Initiate fallback validation}
    \State $ssResult \gets \textsc{SemanticScholarQuery}(q)$
    \State $oaResult \gets \textsc{OpenAlexQuery}(q)$
    \State
    \State \Comment{Cross-validate authors across sources}
    \State $crossRefAuthors \gets \textsc{ExtractAuthors}(bestMatch)$
    \State $ssAuthors \gets \textsc{ExtractAuthors}(ssResult)$
    \State $oaAuthors \gets \textsc{ExtractAuthors}(oaResult)$
    \State
    \State $confirmedAuthors \gets crossRefAuthors \cap ssAuthors \cap oaAuthors$
    \State $suspectAuthors \gets (crossRefAuthors \cup ssAuthors \cup oaAuthors) \setminus confirmedAuthors$
    \State
    \If{$|confirmedAuthors| \geq 2$}
        \State $bestScore \gets bestScore + 10$ \Comment{Multi-source confirmation bonus}
        \State $correctedMetadata \gets \textsc{MergeMetadata}(bestMatch, ssResult, oaResult)$
    \EndIf
    \State
    \If{$|suspectAuthors| > 0$}
        \State $issues \gets issues \cup \{\text{``Potential fabricated authors: ''} + suspectAuthors\}$
    \EndIf
\EndIf

\State $confidence \gets \textsc{ComputeFinalScore}(bestScore, issues)$
\State $apa, bibtex \gets \textsc{GenerateOutputs}(correctedMetadata)$

\State \Return $\{exists: bestScore > 50, confidence, issues, apa, bibtex, sources\}$
\end{algorithmic}
\end{algorithm}
\newpage
\noindent String similarity computation relies on the Levenshtein distance metric, which quantifies the minimum number of single-character edits (insertions, deletions, substitutions) required to transform one string into another. For strings $a$ and $b$, the normalized similarity score is computed as follows:

\begin{equation}
\text{similarity}(a, b) = 1 - \frac{\text{lev}(a, b)}{\max(|a|, |b|)}
\label{eq:levenshtein}
\end{equation}

\noindent where $\text{lev}(a, b)$ denotes the Levenshtein distance and $|a|$, $|b|$ represent string lengths. Prior to comparison, strings undergo normalization through conversion to lowercase and removal of non-alphanumeric characters.

Author matching addresses name format variability through family name extraction and containment checking. The algorithm extracts family names from the primary source author metadata, verifies presence of each family name within the query string, computes author similarity as the proportion of matched authors, and detects potential fabricated authors---capitalized tokens in the query that match neither title words, journal name, year, nor any real author family name. When the fallback mechanism is triggered, the system performs cross-validation by comparing author lists across all queried sources. Authors that appear consistently across CrossRef, Semantic Scholar, and OpenAlex receive higher confidence scores, while authors appearing in only one source or in the query but not in any database are flagged as potentially fabricated. This multi-source confirmation is particularly effective at detecting LLM hallucinations, which frequently insert plausible-sounding but non-existent author names \citep{nicholas2025ecr}.

The composite confidence score integrates component similarities with penalty adjustments and multi-source bonuses. When title similarity exceeds 80\% but author similarity falls below 90\%, the confidence is computed as:

\begin{equation}
\text{confidence} = S_{\text{title}} - 0.5 \times (100 - S_{\text{author}})
\label{eq:confidence1}
\end{equation}

\noindent For structured input with high matching across all fields, the score averages title, author, journal, and year similarities with an additional multi-source confirmation bonus:

\begin{equation}
\text{confidence} = \frac{S_{\text{title}} + S_{\text{author}} + S_{\text{journal}} + S_{\text{year}}}{4} + \beta_{\text{ms}}
\label{eq:confidence2}
\end{equation}

\noindent where $\beta_{\text{ms}} \in [0, 10]$ represents the bonus awarded when author information is validated across multiple databases. Additional penalties are applied for detected issues: title mismatches ($-20$), author mismatches ($-20$), journal discrepancies ($-10$ to $-20$), and each detected fake author ($-10$ to $-20$).\footnote{Threshold and penalty values were empirically calibrated to optimize discrimination between valid references and known hallucinations, with author-related discrepancies weighted more heavily given their stronger diagnostic signal for detecting AI-generated fabrications.}

\section{Features and Usage}

The system provides a unified interface where both quick check and batch check functionality are accessible from a single input area. Users enter references into a common text box, with mode selection determining the processing behavior. This design eliminates the need to navigate between different pages and streamlines the validation workflow. The quick check mode accepts free-form citation text in any standard format (APA, MLA, Chicago, etc.) or informal reference descriptions, with input preprocessing automatically filtering common LaTeX formatting commands to enable researchers to paste references directly from manuscript source files without manual cleaning. This mode is optimized for rapid verification of individual references during manuscript preparation or peer review, returning results within seconds of query submission.

For comprehensive bibliography audits, the batch check mode accepts multiple references either as structured BibTeX entries or as newline-separated citation lists. Upon initiating batch verification, a dedicated results view opens displaying progressive validation outcomes as each reference is processed. The system processes entries sequentially with rate limiting (800ms intervals) to comply with API usage policies, with real-time status updates for each reference.

For each verified reference, the tool generates standardized output in two formats: APA-style citations formatted according to APA 7th edition guidelines suitable for direct manuscript inclusion, and valid BibTeX records with automatically generated citation keys for LaTeX integration. Critically, these outputs derive from authoritative metadata retrieved from the scholarly databases rather than potentially erroneous input, ensuring correctly formatted citations with validated bibliographic details. The batch results interface provides two export options for validated references: a download function that exports all corrected BibTeX entries as a downloadable file ready for import into LaTeX projects or reference managers, and a copy function that transfers all corrected BibTeX entries to the clipboard for immediate pasting. These export features enable researchers to not only verify their bibliographies but also obtain clean, database-validated BibTeX records that can replace potentially erroneous original entries.

\subsection{Comparison with Existing Tools}

Table~\ref{tab:comparison} summarizes the functional differences between the proposed tool and existing reference management systems.

\begin{table}[ht]
\centering
\caption{Functional comparison with existing reference management tools}
\label{tab:comparison}
\begin{tabular}{lccccc}
\toprule
Feature & CheckIfExist & Zotero & Mendeley & EndNote & JabRef \\
\midrule
Reference organization & -- & \checkmark & \checkmark & \checkmark & \checkmark \\
Bibliography formatting & \checkmark & \checkmark & \checkmark & \checkmark & \checkmark \\
Word processor integration & -- & \checkmark & \checkmark & \checkmark & \checkmark \\
Cloud synchronization & -- & \checkmark & \checkmark & \checkmark & -- \\
BibTeX export & \checkmark & \checkmark & \checkmark & \checkmark & \checkmark \\
Immediate validation & \checkmark & -- & -- & -- & -- \\
Hallucination detection & \checkmark & -- & -- & -- & -- \\
Fake author detection & \checkmark & -- & -- & -- & -- \\
Batch verification & \checkmark & -- & -- & -- & -- \\
Multi-source validation & \checkmark & -- & -- & -- & -- \\
Corrected BibTeX output & \checkmark & -- & -- & -- & -- \\
Open source & \checkmark & \checkmark & -- & -- & \checkmark \\
Unlimited free usage & \checkmark & \checkmark & \checkmark & -- & \checkmark \\
\bottomrule
\end{tabular}
\end{table}

As the comparison demonstrates, the proposed system occupies a complementary position within the reference management ecosystem. It does not seek to replace comprehensive tools like Zotero but rather provides a specialized verification capability that these tools lack. Researchers can export their Zotero libraries as BibTeX files, validate them through the proposed interface, identify any problematic entries, and export corrected BibTeX records for reimport. Notably, while commercial hallucination detection services have emerged, many operate under freemium models that restrict the number of verifications available without subscription. The present tool provides unlimited free verification, ensuring accessibility for researchers regardless of institutional resources.

\section{Conclusions}

The proposed system addresses several use cases within contemporary research workflows. Authors can validate reference lists prior to manuscript submission, particularly when bibliographies have been compiled across multiple sources or time periods, with the batch export functionality enabling rapid replacement of uncorrected entries with validated BibTeX records. Researchers using LLMs for literature review assistance can immediately verify any suggested citations before incorporating them into their work, addressing the documented phenomenon of hallucinated references in AI-generated content. Reviewers can efficiently audit citations in manuscripts under review, identifying potential fabrications warranting author clarification---a capability demonstrated to be necessary by the documented presence of hallucinated citations even in papers accepted at top-tier conferences \citep{goldman2026neurips}. Publishers can integrate reference validation into submission pipelines, particularly for manuscripts flagged by AI detection tools, with the multi-source validation providing higher confidence than single-database checks. Researchers conducting large-scale bibliometric analyses can filter potentially spurious references from datasets, ensuring the validity of citation network analyses \citep{taskin2025frozen}.

In conclusion, this paper has presented a practical, open-source solution for immediate bibliographic reference validation through multi-source verification against CrossRef, Semantic Scholar, and OpenAlex databases. While existing tools such as Zotero and Mendeley excel at reference organization and bibliography formatting, they do not address the critical need for citation authenticity verification---a need that has become increasingly urgent with the proliferation of AI-generated content in academic workflows and the documented infiltration of hallucinated citations into peer-reviewed publications at premier venues. By implementing a cascading validation architecture with cross-source author verification and providing instant verification results, the system enables researchers to validate citations at the speed of modern content production. The unified interface, automatic LaTeX command filtering, and BibTeX export functionality streamline integration into existing workflows. The tool complements rather than replaces existing reference management infrastructure, filling a specific gap in the scholarly toolkit. As AI writing assistants become increasingly prevalent, freely accessible tools that support human judgment in maintaining citation integrity become correspondingly important for preserving the reliability of scientific communication.

\section*{Availability}

The tool is released under the MIT License and is available at: \url{https://zabbonat.github.io/References-Validation/}

\bibliographystyle{apalike}
\bibliography{references}

\end{document}